\documentclass[a4,12pt,onecolumn]{IEEEtran}
\IEEEoverridecommandlockouts                    

\usepackage{graphics} 
\usepackage{epsfig} 
\usepackage[]{units} 
\usepackage{amsmath} 
\usepackage{amssymb}  
\usepackage{amsfonts} 
\usepackage{algorithm}
\usepackage{algorithmic}
\usepackage{booktabs}
\usepackage{color}
\usepackage{dblfloatfix}
\usepackage{url}

\ifCLASSOPTIONcompsoc
\usepackage{cite}
\else
\usepackage{cite}
\fi

\DeclareMathOperator*{\argmin}{argmin}   

\newcommand{\real}[1]{\ensuremath{\mathbb{R}^{{#1}}}}

\newcommand{\mat}[1]{\ensuremath{\mathbf{#1}}}
\newcommand{\mats}[2]{\ensuremath{\mathbf{#1}_{#2}}}
\newcommand{\maths}[2]{\ensuremath{\mathbf{\hat{#1}}_{#2}}}

\newcommand{\ve}[1]{\ensuremath{\mathbf{#1}}}
\newcommand{\ves}[2]{\ensuremath{\mathbf{#1}_{#2}}}

\newcommand{\vehs}[2]{\ensuremath{\mathbf{\hat #1}_{#2}}}
\newcommand{\vebs}[2]{\ensuremath{\mathbf{\bar #1}_{#2}}}

\newcommand{\On}[1]{\ensuremath{\mathcal{O} \left({#1}\right)}}
\newcommand{\Exp}[2]{\ensuremath{\mathbb{E}_{#1}\left[#2\right]}}
\newcommand{\Var}[2]{\ensuremath{\mathrm{Var}_{#1}\left[#2\right]}}
\newcommand{\diag}[1]{\ensuremath{\mathrm{diag}\left(#1\right)}}
\newcommand{\tr}[1]{\ensuremath{\mathrm{tr}\left(#1\right)}}
\newcommand{\normaldist}[2]{\ensuremath{\mathcal{N}\left(#1,#2\right)}}

\newcommand{\normtwo}[1]{\ensuremath{\|{#1}\|_2}}
\newcommand{\normfro}[1]{\ensuremath{\|{#1}\|_F}}

\newcommand{\dataset}[1]{\texttt{#1}}

\graphicspath{{figs/}}

\title{Enabling Factor Analysis on Thousand-Subject Neuroimaging Datasets}

\author{\IEEEauthorblockN{Michael J. Anderson\IEEEauthorrefmark{1},  Mihai Capot\u{a}\IEEEauthorrefmark{1},  Javier S. Turek\IEEEauthorrefmark{1}, Xia Zhu\IEEEauthorrefmark{1},Theodore L. Willke\IEEEauthorrefmark{1}, \\%
Yida Wang\IEEEauthorrefmark{5}, Po-Hsuan Chen\IEEEauthorrefmark{2}, Jeremy R. Manning\IEEEauthorrefmark{3}, Peter J. Ramadge\IEEEauthorrefmark{2}, and Kenneth A. Norman\IEEEauthorrefmark{4}}
\IEEEauthorblockA{\IEEEauthorrefmark{1}Parallel Computing Lab, Intel Corporation, Hillsboro, OR\\
\{michael.j.anderson, mihai.capota, javier.turek, xia.zhu, ted.willke\}@intel.com}
\IEEEauthorblockA{\IEEEauthorrefmark{5}Department of Computer Science, Princeton University, Princeton, NJ\\
yida.wang@cs.princeton.edu}
\IEEEauthorblockA{\IEEEauthorrefmark{2}Department of Electrical Engineering, Princeton University, Princeton, NJ\\
\{pohsuan, ramadge\}@princeton.edu}
\IEEEauthorblockA{\IEEEauthorrefmark{3}Department of Psychological and Brain Sciences, Dartmouth College, Hanover, NH\\
jeremy.r.manning@dartmouth.edu}
\IEEEauthorblockA{\IEEEauthorrefmark{4}Department of Psychology and Princeton Neuroscience Institute, Princeton University, Princeton, NJ\\
knorman@princeton.edu}}

\begin{document}

\maketitle
\thispagestyle{empty}
\pagestyle{empty}

\begin{abstract} 
The scale of functional magnetic resonance image data is rapidly increasing as large multi-subject datasets are becoming widely available and high-resolution scanners are adopted. The inherent low-dimensionality of the information in this data has led neuroscientists to consider factor analysis methods to extract and analyze the underlying brain activity. 
In this work, we consider two recent multi-subject factor analysis methods: the Shared Response Model and Hierarchical Topographic Factor Analysis. We perform analytical, algorithmic, and code optimization to enable multi-node parallel implementations to scale. Single-node improvements result in $99\times$ and $1812\times$ speedups on these two methods, and enables the processing of larger datasets. Our distributed implementations show strong scaling of $3.3\times$ and $5.5\times$ respectively with 20 nodes on real datasets. We also demonstrate weak scaling on a synthetic dataset with 1024 subjects, on up to 1024 nodes and 32,768 cores.

\end{abstract}



\section{Introduction}
\label{sec:introduction}



A typical functional magnetic resonance imaging (fMRI) neuroscience experiment consists of scanning a subject that is doing specific tasks while their brain activation is being measured. A sequence of samples in the form of brain volumes is acquired over the duration of the experiment, with a new volume obtained every time of repetition (TR) interval. This results in a few thousands samples of up to a million voxels (volume elements) each. The number of samples that can be gathered from a single subject is limited by time humans can spend in the scanner, so experiments typically span multiple subjects. As more and larger multi-subject datasets are collected and disseminated \cite{ninehundred}, there is a need for new large-scale analysis methods that can leverage multi-subject data.


The spatial and temporal smoothness of fMRI data \cite{huettel2009functional,OpdeBeeck2010smoothness} implies that information is inherently low-dimensional compared to the number of voxels and the number of samples acquired.
Factor analysis is a family of methods that can leverage this characteristic and compute the  statistically shared information across samples \cite{manning2014topographic}.
These methods rely on the assumption that a small number of unobserved latent variables, or \textit{factors}, statistically explain the data. 
In other words, data can be described as a linear combination of these factors plus an additional error (or noise component). 
Factor analysis methods have demonstrated to be practical for neuroscience for interpreting brain activity \cite{Gershman2011tlsa}, predicting cognitive states \cite{chen2015srm,rustamovhyperalignment}, and finding interactions across brain regions \cite{manning2014topographic}. However, applying factor analysis to large multi-subject datasets presents computational challenges. The first is the sheer size of the data; a theoretical 1024-subject dataset collected using high-resolution scanners and two hours of scanning per subject is 17~TB.  Furthermore, the algorithms used to find latent factors  are typically very computationally demanding. 



\begin{figure}
      \centering
      \includegraphics[trim=540 180 470 80,clip,scale=0.13,angle=90,origin=c, width=0.5\textwidth]{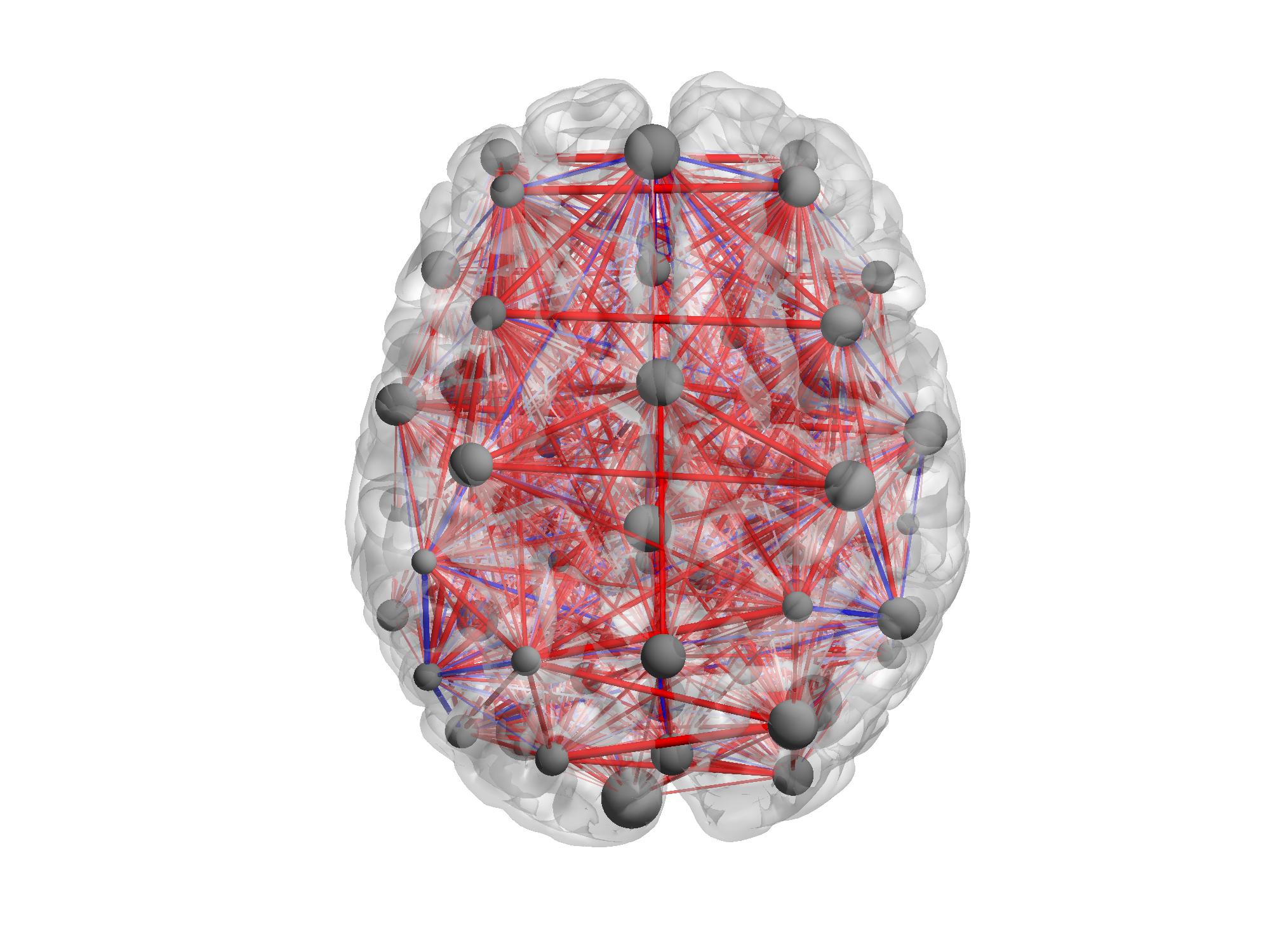} 
      \caption{Brain network discovered by our HTFA implementation.}
      \label{fig:greeneyes_1_network}
\end{figure}

We consider two recent factor analysis techniques designed specifically for multi-subject neuroimaging data: the Shared Response Model (SRM)~\cite{chen2015srm} and Hierarchical Topographic Factor Analysis (HTFA)~\cite{manning2014topographic}. SRM provides a means of aligning the neural activity of subjects through a shared low dimensional response assuming that all subjects have been presented with the same stimuli. The model also can separate multi-subject data into subject-specific neural activity, and shared activity common among all subjects, yielding state-of-the-art predictive results. HTFA provides a means of analyzing functional connectivity between regions of activation in the neuroimaging data, which can either be unique to each subject or follow a global template. HTFA is also useful in interpreting the dynamics of brain activity. We show an example visualization created with our optimized HTFA implementation from real data in Figure~\ref{fig:greeneyes_1_network}.

The algorithm currently used to fit SRM requires a matrix inversion equal to the size of the overall number of voxels for all subjects, which would be an 11-million by 11-million matrix inversion for a 1024-subject high-resolution dataset. 
For HTFA, the core computation is the repeated solving of non-linear least squares on a Jacobian matrix, the size of which depends on the size of a subsampled single subject data and scales quadratically with the number of factors. 
As such, it is currently prohibitive to apply these models to datasets with hundreds of thousands of voxels and hundreds of subjects. In this paper, we address these challenges with (1) linear algebra transformations which reduce computational complexity for both applications, (2) algorithmic improvements to further reduce communication and speed up convergence, and (3) performance optimizations to increase efficiency for distributed and parallel execution. 

Our algorithmic optimizations alone achieve a $671\times$ speedup for SRM and a $42\times$ speedup for HTFA on a single node. Parallel performance optimizations on a single-node provide an additional $2.7\times$ and $2.3\times$ speedup for SRM and HTFA, respectively. We demonstrate the scaling of our implementations on various real and synthetic datasets. For strong scaling on a real 40-subject dataset, our distributed implementations of SRM and HTFA achieve a 5.5$\times$ and $3.3\times$ speedup, respectively, on 20 machines compared to single-node execution. We also weak scale our distributed implementations with a synthetically-generated 1024-subject dataset, simulating two hours worth of high-resolution scanning per subject. We are able to process the entire 17~TB dataset, computing 10 SRM iterations in less than 3 minutes on 16,384 cores, and 1 HTFA iteration in 17 minutes on 32,768 cores, including the time to read the data from a distributed filesystem. 

To summarize, our contributions are the following:
\begin{itemize}
\item We present a new distributed algorithm for SRM with reduced asymptotic complexity and demonstrate a single-node speedup of 1812$\times$ from algorithmic improvements, code optimization, and parallel tuning.
\item We describe algorithmic and implementation improvements for HTFA and demonstrate a single-node speedup of 99$\times$ from algorithmic improvements, code optimization, and parallel tuning.
\item We demonstrate strong scaling for our implementations up to 5.5$\times$ on 20 nodes for real datasets, and successful weak scaling on a 1024-subject synthetic high-resolution dataset, up to 16,384 cores for SRM and 32,768 cores for HTFA. 
\end{itemize}

This work is organized as follows. In Section~\ref{sec:srm}, we describe SRM, analyze its limitations, and propose optimizations to achieve a scalable version. We treat HTFA in a similar manner in Section~\ref{sec:htfa}.  We describe our parallel and distributed implementations of SRM and HTFA in Section~\ref{sec:implementations}.
In Section~\ref{sec:experiments}, we describe our experimental setup for real and synthetic datasets. The results are presented in Section~\ref{sec:results}.  We discuss a series of mathematical and implementation optimizations in Section~\ref{sec:discussion} and conclude with Section~\ref{sec:conclusion}.
\section{Shared Response Model}
\label{sec:srm}

In this section, we describe a recent multi-subject factor analysis method, the Shared Response Model (SRM), and the algorithm used to approximate a solution. SRM has been shown to work well in cases when predictive performance is the primary concern. We describe mathematical transformations and algorithmic optimizations made to the published method which improve the performance of fitting SRM models by reducing asymptotic complexity. We also introduce a distributed algorithm with optimizations which reduce communication between nodes.

\subsection{Model Overview}
A challenge of working with multi-subject datasets is that subjects have distinct anatomical and functional structure and that makes direct aggregation of subject data infeasible. A partial solution to this problem is to apply standard anatomical alignment methods \cite{talairach1988co,fischl1999high,mazziotta2001probabilistic}, but this does not ensure that functional brain topographies align \cite{brett2002,conroy2009fmri,conroy2013inter}.
Recently, functional alignment techniques have appeared that address the misalignment of functional topographies  between subjects \cite{calhoun2001method,friman2001detection,lee2008independent,adali2014diversity,li2009joint,conroy2009fmri,haxby2011common,lorbert2012kernel,rustamovhyperalignment}.
SRM~\cite{chen2015srm} is the state-of-the-art functional alignment method. Given $N$ subjects' volumes synchronously acquired at $T$ time steps, SRM maps functional topographies from every subject to a shared response on a low-dimensional feature space, thus learning an individual mapping for each subject and a shared response across subjects.
The model assumes that a volume $\ves{x}{it} \in \real{V_i}$ with $V_i$ voxels sampled at time $t$ for subject $i$ is the outcome of a random vector:
\begin{equation}
\ves{x}{it} = \mats{W}{i}\ves{s}{t} + \ves{\mu}{i} + \ves{\epsilon}{i}, \qquad \mbox{for all}\quad  i=1\dots N 
\end{equation}
where $\mats{W}{i} \in \real{V_i \times K}$ is the mapping from the $K$-dimensional shared space to
subject $i$ volume space, $\ves{s}{t} \in \real{K}$ is the shared response for all subject volumes at time $t$, $\ves{\mu}{i} \in \real{V_i}$ is the subject specific mean, and \ves{\epsilon}{i} is a 
realization of a \normaldist{\ve{0}}{\rho_i^2\mat{I}} noise vector. 
The model is completed with the assumptions that $\ves{s}{t} \sim \normaldist{\ve{0}}{\mats{\Sigma}{s}}$ with $\mats{\Sigma}{s} \in \real{K\times K}$ and that the mappings \mats{W}{i} are orthogonal, i.e., $\mats{W}{i}^T\mats{W}{i} = \mat{I}$. The orthogonality constraint allows a simple mapping of samples from subject $i$ to subject $j$ by projecting the sample to the feature space and then to the other subject's voxel space, i.e., $\mats{W}{j}\mats{W}{i}^T\ves{x}{it}$.






\subsection{Constrained EM for Model Approximation}
\label{sec:srmlimitations}
The authors in \cite{chen2015srm} propose a constrained Expectation-Maximization (EM) strategy \cite{dempster77em} to approximate the transforms \mats{W}{i} and the shared responses \ves{s}{t} of the $T$ samples.
In the E-step, we compute the sufficient statistics of each sample's shared response:
\begin{eqnarray}
\Exp{\ve{s}|\ve{x}}{\ves{s}{t}} & = & \mats{\Sigma}{s}^T\mat{W}^T(\mat{W}\mats{\Sigma}{s}\mat{W}^T + \mat{\Psi})^{-1}\vehs{x}{t}
\label{eq:srm_mean} \\
\Exp{\ve{s}|\ve{x}}{\ves{s}{t}\ves{s}{t}^T} & = & \Var{\ve{s}|\ve{x}}{\ves{s}{t}} + \Exp{\ve{s}|\ve{x}}{\ves{s}{t}}\Exp{\ve{s}|\ve{x}}{\ves{s}{t}}^T \nonumber \\
&=&\mats{\Sigma}{s} - \mats{\Sigma}{s}^T\mat{W}^T(\mat{W}\mats{\Sigma}{s}^T\mat{W}^T + \mat{\Psi})^{-1}\mat{W}\mats{\Sigma}{s} + \Exp{\ve{s}|\ve{x}}{\ves{s}{t}}\Exp{\ve{s}|\ve{x}}{\ves{s}{t}}^T,
\label{eq:srm_var}
\end{eqnarray}
where $\mat{W} \in \real{V \times K}$ with $V = \sum_{i=1}^{N}{V_i}$ rows representing all the transformations \mats{W}{i} vertically stacked, $\mat{\Psi} \in \real{V \times V}$ is a block diagonal matrix with diagonal $\diag{\rho_1^2\mat{I},\dots,\rho_N^2\mat{I}}$. The vector \vehs{x}{t} represents the demeaned  samples $\vehs{x}{it}=\ves{x}{it}-\ves{\mu}{i}$ at time $t$ vertically stacked for all subjects 
with $\ves{\mu}{i} = \frac{1}{T}\sum_{t=1}^{T}\ves{x}{it}$. 
Note that the estimated shared response is obtained as $\vehs{s}{t} = \Exp{\ve{s}|\ve{x}}{\ves{s}{t}}$.

Once we obtain \Exp{\ve{s}|\ve{x}}{\ves{s}{t}} and \Exp{\ve{s}|\ve{x}}{\ves{s}{t}\ves{s}{t}^T} in the E-step, we compute the M-step that updates the hyperparameters \mats{W}{i}, $\rho_i^2$, and \mats{\Sigma}{s} as follows:
\begin{eqnarray}
\mats{A}{i} & = & \frac{1}{2}\sum_{t} \vehs{x}{it}  \Exp{\ve{s}|\ve{x}}{\ves{s}{t}}^T \nonumber \\
\mats{W}{i}^{new} & = & \mats{A}{i}\left(\mats{A}{i}^T\mats{A}{i}\right)^{-\frac{1}{2}} \label{eq:srm_mapping_update} \\ 
\rho_i^{2,new} & = & \frac{1}{T\cdot V}\sum_t\|\vehs{x}{it}\|_2^2 + \frac{1}{T\cdot V}\sum_t \tr{\Exp{\ve{s}|\ve{x}}{\ves{s}{t}\ves{s}{t}^T}}  -2\sum_t \vehs{x}{it}^T \mats{W}{i}^{new}\Exp{\ve{s}|\ve{x}}{\ves{s}{t}}  \label{eq:srm_rho_update}  \\
\mats{\Sigma}{s}^{new} & = & \frac{1}{T}\sum_t \Exp{\ve{s}|\ve{x}}{\ves{s}{t}\ves{s}{t}^T}. \label{eq:srm_variance_update}
\end{eqnarray}
This method for approximating SRM iteratively computes the E-step and M-step until some stopping criterion is met.

To understand the model's computational limitations we first note that the model size depends on $T$ samples (TRs), $N$ subjects and their $V_i$ voxels, 
and the number of features $K$. The latter value  determines the cost of some computations and
the memory needed to store the mappings \mats{W}{i} and the shared responses \ves{s}{t}.
The assumption that the shared information has a low rank dictates that $K$ should be small. It was shown that the best accuracy is obtained when the parameter $K$ is set to a value  ranging from tenths to hundreds of features~\cite{chen2015srm}. The importance of $K$ being small will be evident below.

The EM algorithm described in Equations \eqref{eq:srm_mean}--\eqref{eq:srm_variance_update} has several advantages.
Equations \eqref{eq:srm_mapping_update} and \eqref{eq:srm_rho_update} in the M-step can be computed in parallel across subjects. Equation \eqref{eq:srm_variance_update} is a summation over $T$ small matrices of size $K$ by $K$, which also does not pose any challenge. In contrast, the E-step requires inversion of $\mat{\Phi} = \mat{W}\mats{\Sigma}{s}\mat{W}^T + \mat{\Psi}$, 
a matrix of size $V$ by $V$ elements depending on the number of subjects $N$ and the numbers of voxels per subject $V_i$. Computing and storing 
\mat{\Phi} and its inverse $\mat{\Phi}^{-1}$ is very challenging and sometimes may be nonviable. Because \mat{\Phi} is symmetric, Cholesky factorization is the most stable method for computing its inverse. While \mat{\Phi} requires \On{V^2} memory, the Cholesky factorization requires additional \On{V^2} memory and has a \On{V^3}  runtime  complexity. For instance, when $N=10$ subjects with $V_i=50,000$ voxels each, the size of the matrix would be 500,000 by 500,000 and would require 20~GB for storing \mat{\Phi} only, even before computing the Cholesky factorization.

\subsection{Reducing the Inversion}
\label{sec:srmoptimizations}

As discussed in Section \ref{sec:srmlimitations}, the runtime and memory bottleneck is given by
the inversion of the big matrix \mat{\Phi}. We derive an analytical alternative formula to avoid computing \mat{\Phi}, and hence reducing the computational runtime and memory usage. First, we apply the matrix inversion lemma to the first two terms in Equation \eqref{eq:srm_var}:
\begin{equation}
\mats{\Sigma}{s} - \mats{\Sigma}{s}^T\mat{W}^T\mat{\Phi}^{-1}\mat{W}\mats{\Sigma}{s}  = \left(\mats{\Sigma}{s}^{-1} + \mat{W}^T\mat{\Psi}^{-1}\mat{W}\right)^{-1}. \label{eq:srm_inversion_opt}
\end{equation}
While Equation \eqref{eq:srm_var} requires inverting a $V^2$ matrix, Equation \eqref{eq:srm_inversion_opt} supplants that by computing three much smaller inversions. The matrices \mats{\Sigma}{s}  and $\mats{\Sigma}{s}^{-1} + \mat{W}^T\mat{\Psi}^{-1}\mat{W}$ have a very small size, $K$ by $K$, and their inversions can be computed fast with a Cholesky factorization on one machine. The matrix \mat{\Psi} is a diagonal matrix and its inversion is computed by inverting its diagonal elements. Based on the orthogonality property of \mats{W}{i}, i.e., $\mats{W}{i}^T\mats{W}{i}=\mat{I}$, we note that
\begin{equation}
\mat{W}^T\mat{\Psi}^{-1}\mat{W} = \sum_{i}\mats{W}{i}^T \cdot \rho_i^{-2}\mat{I} \cdot \mats{W}{i} = \rho_0\mat{I}, \label{eq:srm_wt_psi_w}
\end{equation}
where $\rho_0 = \sum_{i=1}^{N} \rho_i^{-2}$.
Equation \eqref{eq:srm_wt_psi_w} helps in further reducing the number of matrix multiplications for computing \eqref{eq:srm_var}.

Now, we derive an alternative formulation for \Exp{\ve{s}|\ve{x}}{\ves{s}{t}} in Equation \eqref{eq:srm_mean},  by applying the matrix inversion lemma as in \eqref{eq:srm_inversion_opt} and some additional linear algebra steps:

\begin{equation}
\mats{\Sigma}{s}^T\mat{W}^T(\mat{W}\mats{\Sigma}{s}\mat{W}^T + \mat{\Psi})^{-1}   
= \mats{\Sigma}{s}^T\left[\mat{I} - \rho_0 \left(\mats{\Sigma}{s}^{-1} + \rho_0\mat{I}\right)^{-1} \right] \mat{W}^T\mat{\Psi}^{-1}. \label{eq:srm_mean_opt}
\end{equation}
The result above requires the same three inversions needed in Equation \eqref{eq:srm_inversion_opt}.
Therefore, the simplified computations in the E-step require only to invert dense matrices of $K$ by $K$ elements or a diagonal matrix with $N$ different values. Hence, the memory complexity is reduced from \On{V^2} to \On{K^2}, much less than originally needed to invert \mat{\Phi}. It is worth noting that the required memory for these computations depends only on the number of features $K$ and is independent of the number of aggregated voxels $V$ in the data (where $K \ll V$). 

\subsection{Reducing Communication for Distributed Computation}

Recall that computing the M-step is embarrassingly parallel assuming data is distributed by subject.
However, the E-step requires centralized computation of the shared responses.
Therefore, we study next the communications needed for a distributed version of the EM algorithm. 
Plugging  Equation \eqref{eq:srm_mean_opt} into Equation \eqref{eq:srm_mean} and rewriting it to partition by subject samples, we obtain:
\begin{equation}
\Exp{\ve{s}|\ve{x}}{\ves{s}{t}} = \mats{\Sigma}{s}^T\left[\mat{I} - \rho_0 \left(\mats{\Sigma}{s}^{-1} + \rho_0\mat{I}\right)^{-1} \right] \sum_{i}\rho_i^{-2}\mats{W}{i}^T \vehs{x}{it}. \label{eq:srm_mean_distributed} 
\end{equation}
This equation allows us to identify what needs to be transferred to compute the shared responses in the E-step.
To compute \eqref{eq:srm_mean_distributed}, we should transfer each $\rho_i^2$  to a master process and reduce the summation $\sum_{i}\rho_i^{-2}\mats{W}{i}^T \vehs{x}{it}$. Each transferred matrix (size $K\times T$) is relatively small. 
The resulting \Exp{\ve{s}|\ve{x}}{\ves{s}{t}} (a $K\times T$ matrix) should be broadcast to all processes for the M-step computation.
The matrices \Exp{\ve{s}|\ve{x}}{\ves{s}{t}\ves{s}{t}^T} from \eqref{eq:srm_var} in the E-step are used for updating $\mats{\Sigma}{s}^{new}$ in \eqref{eq:srm_variance_update} and computing the trace for $\rho_i^{2,new}$ in \eqref{eq:srm_rho_update}. Although broadcasting $T$ of such $K \times K$ matrices is not limiting, we suggest avoiding it by updating $\mats{\Sigma}{s}^{new}$ in the master process by noting that
\begin{equation}
\mats{\Sigma}{s}^{new} = \left(\mats{\Sigma}{s}^{-1} + \rho_0 \mat{I}\right)^{-1} + 
\frac{1}{T} \sum_t  \Exp{\ve{s}|\ve{x}}{\ves{s}{t}}\Exp{\ve{s}|\ve{x}}{\ves{s}{t}}^T,
\label{eq:srm_variance_update_optimized}
\end{equation}
and that $ \nicefrac{1}{T}\sum_t \tr{\Exp{\ve{s}|\ve{x}}{\ves{s}{t}\ves{s}{t}^T}} = $ $\tr{\nicefrac{1}{T}\sum_t \Exp{\ve{s}|\ve{x}}{\ves{s}{t}\ves{s}{t}^T}}$ $ = \tr{\mats{\Sigma}{s}^{new}}$. Therefore, we first compute $\mats{\Sigma}{s}^{new}$ in the master process and then broadcast its trace. This reduces the amount of communication from \On{TK^2} to \On{TK}. 
The distributed EM algorithm for SRM is summarized in Algorithm \ref{alg:srm}.

\begin{algorithm}
\caption{Distributed Expectation-Maximization for Shared Response Model (D-SRM)}
\label{alg:srm}
\begin{algorithmic}[1]
\REQUIRE A set of $T$ samples \ves{x}{it} for each subject, and the number of features $K$.
\ENSURE A set of $N$ mappings \mats{W}{i}, the shared responses \ves{s}{t}
\STATE \textbf{Initialization:} In each process, initialize \mats{W}{i} to a random orthogonal matrix, $\rho_i^2 = 1$, and demean the input by $\vehs{x}{it} = \ves{x}{it} - \ves{\mu}{i}$ with $\ves{\mu}{i} = \frac{1}{T}\sum_{t=1}^{T}\ves{x}{it}$.
\WHILE{stopping criterion not met} 
\STATE \textit{E-Step}:
\STATE Reduce $\sum_{i}\rho_i^{-2}\mats{W}{i}^T \vehs{x}{it}$ and transfer $\rho_i^2$ from each process to the master process.
\STATE Compute \Exp{\ve{s}|\ve{x}}{\ves{s}{t}} in \eqref{eq:srm_mean_distributed} in the master process.
\STATE \textit{M-Step}:
\STATE Update $\mats{\Sigma}{s}$ in \eqref{eq:srm_variance_update_optimized} and its trace in the master process.
\STATE Broadcast \Exp{\ve{s}|\ve{x}}{\ves{s}{t}} and the trace of $\mats{\Sigma}{s}$ to all processes.
\STATE In each process, update $\mats{W}{i}$ and $\rho_i^2$ following \eqref{eq:srm_mapping_update} and \eqref{eq:srm_rho_update}.
\ENDWHILE
\end{algorithmic}
\end{algorithm}

\section{Hierarchical Topographic Factor Analysis}
\label{sec:htfa}

In this section, we discuss Hierarchical Topographic Factor Analysis (HTFA), an interpretable model that can be used by neuroscientists analyze and visualize brain networks. As we did with the SRM, we apply both mathematical transformations and algorithmic optimizations to improve the performance of HTFA compared to the baseline published algorithm. We also present a scalable distributed implementation. 

\subsection{Model Overview}
A cognitive state can be interpreted as a network of nodes and edges, representing regions of activity in the brain and their correlated activation. 
Topographic factor analysis (TFA)~\cite{manning2014topographic} discovers these activity regions and their connectivity pattern through time. 
This method is based on a Bayesian model that describes an fMRI volume as a linear combination of factors, which are assumed to be Gaussian distributions. Each such distribution depicts a three-dimensional sphere representing an activity region. 
For a given series of brain volumes, TFA uses variational inference to estimate the centers and widths of the factors and the linear combination weights.

HTFA~\cite{manning2014htfa} is a multi-subject extension of TFA that further assumes that all subjects exhibit a global template of activity regions. Thus, every subject is described as a small perturbation of this global template. 
Let $\mats{X}{i} \in \real{T_i \times V_i}$ represent subject $i$'s data as a matrix with $T_i$ fMRI samples of $V_i$ voxels each and vectorized in \mats{X}{i}'s rows.
Then, each subject is approximated with a factor analysis model
\begin{equation}
\mats{X}{i} = \mats{W}{i} \mats{F}{i} + \mats{E}{i}, \label{eq:htfa_factor_analysis}
\end{equation}
where $\mats{E}{i} \in \real{T_i \times V_i}$ is an error term, $\mats{W}{i} \in \real{T_i \times K}$ are the weights of the factors in $\mats{F}{i} \in \real{K \times V_i}$. The factor (row) $k$ in \mats{F}{i} represents a ``sphere'' or normal distribution with center \ves{\mu}{i,k} and width $\lambda_{i,k}$ located relatively to the positions of the voxels in the volume. 

\begin{figure}
      \centering
      \includegraphics[trim=200 145 250 160,clip,scale=0.5,width=0.5\textwidth]{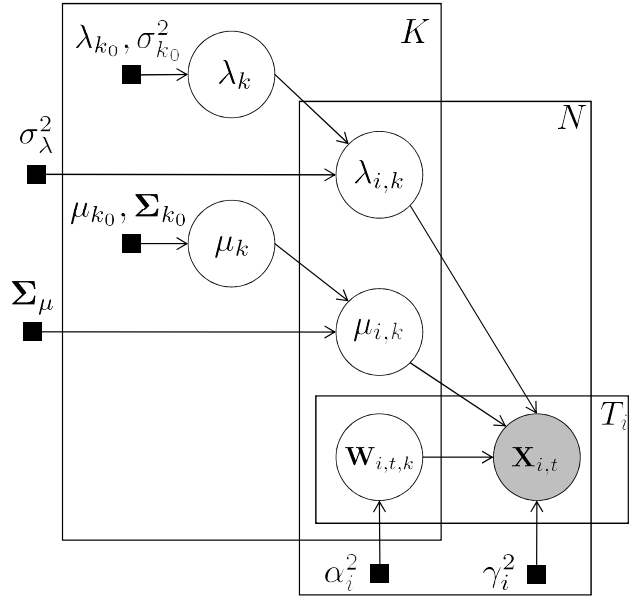} 
      \caption{Graphical model representation of HTFA. White nodes are latent variables, gray nodes represent observations, and the black squares are the hyperparameters of the model. The model has $K$ factors, $N$ subjects and $T_i$ samples each.} 
      \label{fig:model_htfa}
\end{figure}

HTFA defines the local factors in \mats{F}{i} as perturbations of the factors of a global template in \mat{F}. 
Therefore, the factor centers $\ves{\mu}{i,k}$  for all subjects are obtained from a normal distribution with mean \ves{\mu}{k} and covariance \mats{\Sigma}{\mu}. The mean \ves{\mu}{k} represents the center of the global $k^{th}$ factor, while \mats{\Sigma}{\mu} determines the distribution of the possible distance between the global and the local center of the factor. Similarly, the widths $\lambda_{i,k}$  for all subjects are drawn from a normal distribution with mean $\lambda_k$, the width of the global $k^{th}$ factor, and variance $\sigma_\lambda^2$. The model assumes that \mats{\Sigma}{\mu} and $\sigma_\lambda^2$ are constants and the same for all factors. On top of the global parameters \ves{\mu}{k} and $\lambda_k$, the model defines Gaussian priors for each, respectively, \normaldist{\ves{\mu}{k_0}}{\mats{\Sigma}{k_0}} and \normaldist{\lambda_{k_0}}{\sigma_{k_0}^2}. 
In addition, the columns of the weight  matrices \mats{W}{i} are modeled with a \normaldist{\ve{0}}{\alpha_i^2\mat{I}} distribution and  the elements in the noise term \mats{E}{i} are assumed to be independent with a  \normaldist{0}{\gamma_{i}^2} distribution. The associated graphical model is shown in Figure~\ref{fig:model_htfa}.

\subsection{Approximating the MAP Estimator} 
\label{sec:htfamap}

Based on this hierarchical model, the maximum a-posteriori probability (MAP) estimator can find the centers and widths of the distributions. However, the exact estimator is hard to compute. Hence, Manning et al.~\cite{manning2014topographic} suggested a black box variational inference technique to achieve an approximate estimate.
The method consists of a global and a local step that iteratively update the parameters. The global step updates the  parameters of the $K$ distributions in the global template. The local step updates for each subject $i$ the weight matrices \mats{W}{i}, the local centers \ves{\mu}{i,k} and widths $\lambda_{i,k}$.
To update the parameters of the factors in \mats{F}{i}, the local step solves 
\begin{eqnarray}
\left\{\vehs{\mu}{i,k},\hat{\lambda}_{i,k} \right\}_{k} &= 
\argmin_{\left\{\ves{\mu}{i,k},\lambda_{i,k} \right\}_k} & \frac{1}{2\sigma_i^2}\normfro{\mats{X}{i} - \mats{W}{i}\mats{F}{i}}^2 
+\frac{1}{2\phi_i} \sum_{k=1}^{K} \left(\ves{\mu}{i,k} - \vehs{\mu}{k}\right)\mats{\Sigma}{\mu}^{-1} \left(\ves{\mu}{i,k} - \vehs{\mu}{k}\right)^T 
\nonumber \\
&&+ \frac{1}{2\phi_i \sigma_{\lambda}^2 }\sum_{k=1}^{K} \left(\lambda_{i,k} - \hat{\lambda}_{k} \right)^2, 
\label{eq:htfa_subject_update_rbfs}
\end{eqnarray}
where $\phi_i$ is a subsampling coefficient defined below. Each row of $\mats{F}{i}$ is a factor obtained by evaluating the radial basis function (RBF) with center at \ves{\mu}{i,k} and width $\lambda_{i,k}$
\setcounter{equation}{13}
\begin{equation}
f_{i,k}\left(\ve{p};\ves{\mu}{i,k},\lambda_{i,k}\right) = \exp \left\{-\frac{ \normtwo{\ve{p}-\ves{\mu}{i,k}}^2}{\lambda_{i,k}} \right\},
\label{eq:htfa_factor}
\end{equation}
in positions $\ve{p} \in \real{3}$ for all the voxels in the three-dimensional voxel space of the brain.

The objective function in~\eqref{eq:htfa_subject_update_rbfs} is non-linear because of the definition of the factors in \mats{F}{i}. Therefore, Manning et al.~\cite{manning2014topographic} propose to decompose the objective function as a sum of subfunctions and compute a solution using an unconstrained non-linear least squares (NLLS) solver, implemented with a trust-region reflective method \cite{Coleman1996trf}. This is an iterative method that computes the Jacobian matrix of partial derivatives at each iteration. Hence, the size of the Jacobian matrix depends on the number of variables and the number of subfunctions defined in the problem. 
In the case of HTFA, there are $4K$ variables: $K$ factors with a three dimensional center \ves{\mu}{i,k} and a scalar width $\lambda_{i,k}$. The first term of the objective function containing the Frobenius norm can be described as a summation over $V_iT_i$ subfunctions, each given by a squared element of the matrix $\left(\mats{X}{i} - \mats{W}{i}\mats{F}{i}\right)$.
The second and third terms in \eqref{eq:htfa_subject_update_rbfs} can be partitioned in $K$ subfunctions each. Summing up, there are $V_iT_i + 2K$ functions, and the size of the Jacobian matrix is $4K$ by $V_iT_i + 2K$. Although the Jacobian matrix is computed per subject and does not depend on the number of subjects $N$, it still can be prohibitively large. For example, solving the problem for a subject with $V_i=50,000$ voxels, $T_i=2,000$ samples and for $K=100$ factors, this matrix will occupy $\sim$300~GB of memory. Manning et al.~\cite{manning2014topographic} mitigate this limitation by randomly subsampling the rows (TRs) and columns (voxels) from the data matrix \mats{X}{i}, reducing it to $\tilde{T}_i$ samples by $\tilde{V}_i$ voxels ($\tilde{T}_i<T_i$ and $\tilde{V}_i<V_i$). This requires adding the sampling coefficient $\phi_i=\nicefrac{T_iV_i}{\tilde{T}_i\tilde{V}_i}$ to compensate between the weights of the first and other terms in \eqref{eq:htfa_subject_update_rbfs}.

After the subject centers and widths are updated via \eqref{eq:htfa_subject_update_rbfs}, we update the weights matrix \mats{W}{i} for each subject by solving
\begin{equation}
\maths{W}{i} ={\arg \min}_{\mats{W}{i}} \normfro{\mats{X}{i} - \mats{W}{i}\mats{F}{i}}^2 + \frac{1}{\alpha_i^2}\normfro{\mats{W}{i}}^2. 
\label{eq:htfa_subject_update_weights}
\end{equation}
Problem~\eqref{eq:htfa_subject_update_weights} admits a closed-form solution of the form of ridge regression, i.e., $\maths{W}{i} = \mats{X}{i}\mats{F}{i}^T\left(\mats{F}{i}\mats{F}{i}^T + \nicefrac{1}{\alpha_i^2} \mat{I}\right)^{-1}$. Due to the subsampling of the data matrix \mats{X}{i} employed to update the factors of subject $i$, only a subset of the rows in \maths{W}{i} are updated.
The weights and local parameters are updated solving \eqref{eq:htfa_subject_update_rbfs} and \eqref{eq:htfa_subject_update_weights}, alternately, until a convergence criterion is met.

Then, the hyperparameters of the global template priors are updated given the local estimates and under the assumption that the posterior has a conjugate prior with normal distribution. This yields the following formulas for the updating the hyperparameters \ves{\mu}{k}, \mats{\Sigma}{k}, ${\lambda}_{k}$, and ${\sigma}_{k}^2$:
\begin{align}
\vehs{\mu}{k}^{new} & = \left( \maths{\Sigma}{k}^{-1} + N \mats{\Sigma}{\mu}^{-1} \right)^{-1} \left( \maths{\Sigma}{k}^{-1}\vehs{\mu}{k} + N \mats{\Sigma}{\mu}^{-1} \vebs{\mu}{k} \right) \label{eq_htfacentermean} \\
\maths{\Sigma}{k}^{new} & =  \left( \maths{\Sigma}{k}^{-1} + N \mats{\Sigma}{\mu}^{-1} \right)^{-1} \label{eq_htfacentercov} \\
\hat{\lambda}_{k}^{new} & = \left( \hat{\sigma}_{k}^{-2} + N {\sigma}_{\lambda}^{-2} \right)^{-1} \left( \hat{\sigma}_{k}^{-2}\hat{\lambda}_{k} + N {\sigma}_{\lambda}^{-2} \bar{\lambda}_{k} \right) \label{eq_htfawidthmean} \\
\hat{\sigma}_{k}^{2,new} & = \left( \hat{\sigma}_{k}^{-2} + N {\sigma}_{\lambda}^{-2} \right)^{-1}, \label{eq_htfawidthvar}
\end{align}
where $\vebs{\mu}{k} = \nicefrac{1}{N}\sum_{i} \vehs{\mu}{i,k}$, $\bar{\lambda}_{k} = \nicefrac{1}{N}\sum_{i} \hat{\lambda}_{i,k}$. The above parameters \vehs{\mu}{k}, \maths{\Sigma}{k}, $\hat{\lambda}_{k}$, and $\hat{\sigma}_{k}^{2}$ are initialized to \ves{\mu}{k0}, \mats{\Sigma}{k0}, $\lambda_{k0}$, and $\sigma_{k0}^2$, respectively.
The parameters \mats{\Sigma}{\ve{\mu}}, and ${\sigma}_{\lambda}^2$ are computed during initialization and remain constant through the iterations.

The local update step that solves \eqref{eq:htfa_subject_update_rbfs} and \eqref{eq:htfa_subject_update_weights} is embarrassingly parallel across subjects. 
Nevertheless, the NLLS solver is a runtime bottleneck because of its dependence on the big Jacobian matrix for the solver computations and because the problem is solved several times in each local update step. 
As the centers and widths of the local factors change when solving \eqref{eq:htfa_subject_update_rbfs}, the factor matrix \mats{F}{i} is re-computed whenever the cost function in \eqref{eq:htfa_subject_update_rbfs} is evaluated inside each iteration of the NLLS solver, adding to the runtime.
Despite subsampling the data matrix \mats{X}{i}, in terms of memory, the Jacobian matrix size dictates the complexity of the algorithm.
In the following section, we describe optimizations targeting mainly the local updates step which (1) reduce the size of the Jacobian matrix, (2) use problem-specific assumptions to simplify the problem,  (3) optimize the factors calculation, and (4) reduce the number of matrix inversions.



\subsection{Optimizations to the MAP estimator}
\label{sec:htfaoptimizations}

The computation complexity of NLLS solver makes it a direct target for optimization. We recall that the size of Jacobian matrix depends on the number of variables and the number of subfunctions. It is possible to partition the Frobenius norm term in \eqref{eq:htfa_subject_update_rbfs} with a different subfunction granularity to reduce the number of subfunctions and hence, the size of the Jacobian matrix. However, the runtime does not necessarily reduce, as recomputing the Jacobian matrix in internal iterations of the NLLS solver requires the same number of operations independent of the granularity.
Instead, we consider an alternative approach. We partition the variables into two blocks: (a) the centers \ves{\mu}{i,k} and (b) the widths $\lambda_{i,k}$. We fix the values of the variables in block (b) and solve~\eqref{eq:htfa_subject_update_rbfs} only for the block (a). Then, we do the opposite and fix the variables in block (a) and solve for those in block (b). 
Although this block partitioning requires us to solve two subproblems with an NLLS solver, the computation complexity of the Jacobian matrix is smaller. The reasons are that the number of variables is reduced to the block size and $K$ subfunctions can be dropped as they are constant when fixing any of the blocks. 
In addition, this has the potential to reduce the number of steps to converge in the local updates, resembling a block coordinate descent approach.

The original algorithmic implementation utilizes an unconstrained NLLS solver. Since the model assumes that the location of each factor's center is within the human brain, 
we add constraints to~\eqref{eq:htfa_subject_update_rbfs}: the centers \ves{\mu}{i,k} are bound to be within the brain volume and the value of widths $\lambda_{i,k}$ are at least 4\% of the brain diameter and at most 180\% of it; these two parameters can be tuned for different datasets.
This modification to the problem has the potential to speed-up the solution by restricting the feasible region and maintaining the centers within the brain area in intermediate solutions of the NLLS solver. The unconstrained version may yield negative or zero width values in some iterations, leading to positive exponents in the RBF evaluation and hence to large values that might cause numerical overflow issues in the NLLS solver. 
Moreover, the contrained implementation may yield better solutions by avoiding factors of small sizes that represent a single noisy voxel. A solution to the constrained version of~\eqref{eq:htfa_subject_update_rbfs} can be obtained using a constrained NLLS. We opt for a trust-region reflective solver.

During the local factor and weight updates, each factor is evaluated to build the updated matrix \mats{F}{i} explicitly. For this purpose, the value of the function $f_{i,k}(\ve{p};\ves{\mu}{i,k},\lambda_{i,k})$ is calculated for all the voxel locations $\ve{p}=(x,y,z)$. In particular, $\ve{p}=(x,y,z)$ represents the positions of the voxels in a three-dimensional grid. Because of the bounded size of the brain and the spatial contiguity of the voxels, the number of unique elements in each dimension ($x,y,z$) for all locations \ve{p} is much lower than the number of voxels.
In addition, the Euclidean distance term in the exponent of the RBF can be expressed as 
$\left(\ves{p}{x}-\ves{\mu}{x}\right)^2 + \left(\ves{p}{y}-\ves{\mu}{y}\right)^2 + \left(\ves{p}{z}-\ves{\mu}{z}\right)^2$. Therefore, we can reduce the number of subtractions and multiplications (i.e., the squares) for computing \mats{F}{i} by caching the values $\left(\ves{p}{coor}-\ves{\mu}{coor}\right)^2$ for each coordinate $x$, $y$, or $z$ in a look up table. Assuming that the voxel locations are inside a cube of size $n_x \times n_y \times n_z$, the effect of caching  reduces  from \On{n_x n_y n_z} floating point subtractions and multiplications  to \On{n_x + n_y + n_z} for each factor.



The global posterior updates in Equations~\eqref{eq_htfacentermean}--\eqref{eq_htfawidthvar} require three matrix inversions per factor. Applying the matrix inversion lemma on these equations we obtain 
\begin{align}
\vehs{\mu}{k}^{new} &  = A\frac{\mats{\Sigma}{\mu}}{N} \vehs{\mu}{k} 
	+ \maths{\Sigma}{k}A\vebs{\mu}{k}  \label{eq_newhtfacentermean} \\ 
\maths{\Sigma}{k}^{new} & =  \maths{\Sigma}{k} A\frac{\mats{\Sigma}{\mu}}{N} \label{eq_newhtfacentercov} \\
\hat{\lambda}_{k}^{new} & = 
 b\frac{{\sigma}_{\lambda}^2}{N} \hat{\lambda}_{k}  + \hat{\sigma}_{k}^2b\bar{\lambda}_{k} \label{eq_newhtfawidthmean}\\
\hat{\sigma}_{k}^{2~new} & = \hat{\sigma}_{k}^2b\frac{{\sigma}_{\lambda}^2}{N},  \label{eq_newhtfawidthvar}
\end{align}
where $A  = \left( \maths{\Sigma}{k} + \nicefrac{\mats{\Sigma}{\mu}}{N}  \right)^{-1}$, and $b = \left(\hat{\sigma}_{k}^2  + \nicefrac{{\sigma}_{\lambda}^2}{N}  \right)^{-1}$.
In this manner, we reduce the number of inversions per factor in the global update step.
The optimized version of the method is summarized in Algorithm~\ref{alg:htfa}.

\begin{algorithm}
\caption{Distributed MAP estimator for HTFA (HTFA D-MAP)}
\label{alg:htfa}
\begin{algorithmic}[1]
\REQUIRE A set samples \mats{X}{i} for each subject, the number of features $K$.
\ENSURE The $K$ estimated global template centers $\vehs{\mu}{k}$ and widths $\hat{\lambda}_{k}$, $N$ sets of $K$ estimated local centers $\vehs{\mu}{i,k}$ and widths $\hat{\lambda}_{i,k}$, and 
$N$ weights matrices \mats{W}{i}.  
\STATE \textbf{Initialization:} Initialize the global prior parameters \ves{\mu}{k_0}, \mats{\Sigma}{k_0}, $\lambda_{k_0}$, and ${\sigma}_{k_0}^2$ as in  \cite{manning2014topographic} using one subject.
\STATE Set $\vehs{\mu}{k}=\ves{\mu}{k0}$, $\maths{\Sigma}{k}=\mats{\Sigma}{k0}$, $\hat{\lambda}_{k}=\lambda_{k0}$, $\hat{\sigma}_{k}^{2}=\sigma_{k0}^2$, $\mats{\Sigma}{\ve{\mu}}=\mats{\Sigma}{k0}$, and $\sigma_{\lambda}^{2}=\sigma_{k0}^2$.
\WHILE{stopping criterion not met} 
\STATE Broadcast global centers \vehs{\mu}{k} and widths $\hat{\lambda}_{k}$ to all processes.
\STATE \textit{Local update (on each process):}
\STATE Set local prior to latest global prior $\vehs{\mu}{i,k} = \vehs{\mu}{k}$ and $\hat{\lambda}_{i,k} = \hat{\lambda}_{k}$. 
\WHILE{local stopping criterion not met} 
\STATE Subsample rows and columns of \mats{X}{i}
\STATE Compute factors matrix \mats{F}{i} and update \maths{W}{i} by solving Problem~\eqref{eq:htfa_subject_update_weights} with ridge regression.
\STATE Update subject centers $\vehs{\mu}{i,k}$ solving Problem \eqref{eq:htfa_subject_update_rbfs} with a constrained NLLS solver with subject widths $\hat{\lambda}_{i,k}$ fixed.
\STATE Update subject widths $\hat{\lambda}_{i,k}$ solving Problem \eqref{eq:htfa_subject_update_rbfs} with a constrained NLLS solver with subject centers $\vehs{\mu}{i,k}$ fixed.
\ENDWHILE
\STATE Gather the local centers $\vehs{\mu}{i,k}$ and widths $\hat{\lambda}_{i,k}$ from each process.
\STATE \textit{Global template update (on the master process):}
\STATE Update the global centers $\vehs{\mu}{k}$, widths $\hat{\lambda}_{k}$  and the hyperparameters \maths{\Sigma}{k} and $\hat{\sigma}_{k}^2$ using Equations \eqref{eq_newhtfacentermean}--\eqref{eq_newhtfawidthvar}.
\ENDWHILE
\STATE Update each subject's weights matrix $\ves{W}{i}$ in each process.
\end{algorithmic}
\end{algorithm}

\begin{figure*}
      \centering
      \includegraphics[width=\linewidth]{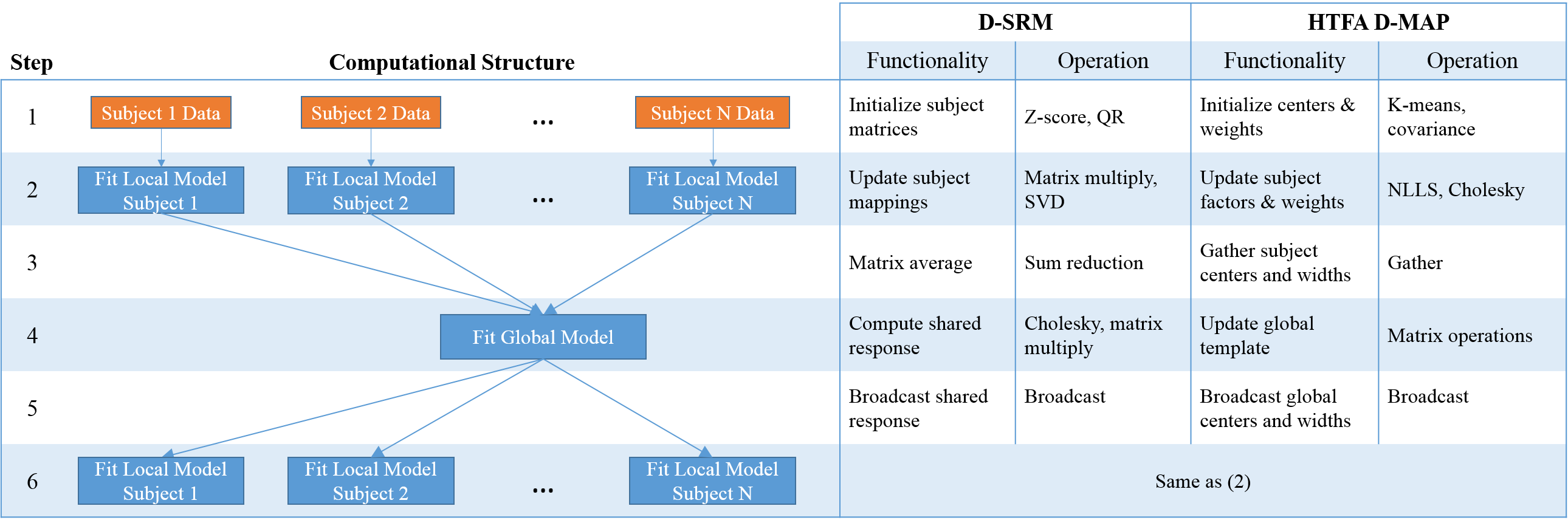} 
      \caption{Computational structure of D-SRM and HTFA D-MAP, and the main operations involved at each step.} 
      \label{fig:implementation}
\end{figure*}

\section{Parallel and Distributed Implementation}
\label{sec:implementations}

Our parallel implementations, D-SRM and HTFA D-MAP, both follow a similar map-reduce structure, which is shown in Figure~\ref{fig:implementation}. Both algorithms start by fitting local models in parallel for each subject (Step 2), then combining the local models with a gather or reduction (Step 3), and fitting a global model for all subjects (Step 4). The global model is then broadcast (Step 5) and the local models are re-fit using the updated global information. This process continues in an iterative manner for both algorithms. The columns on the right of the figure show a summary of the main computations, at each step, for both D-SRM and HTFA D-MAP. The following subsections provide further details of our parallel and distributed implementation and specific code optimizations.

\subsection{Single-node Implementation}

Our baseline SRM and HTFA implementations are written in Python and use the NumPy and SciPy libraries for matrix representations and operations. We use the Anaconda distribution of Python, which recently added support for using the Intel Math Kernel Library (MKL) to accelerate NumPy and SciPy in time-consuming math operations such as matrix multiplication and matrix inversion. Our new optimized implementations continue to use Python for compatibility with other neuroscience tools which in turn use Python for productivity.

Some operations in NumPy and SciPy, such as matrix multiplication, run very efficiently on all cores automatically using Anaconda's MKL-backed Numpy. However, we found that certain operations were running serially. These serial functions became the bottleneck in certain parallel configurations, so we wrote small OpenMP-parallelized NumPy extension modules in C++ to speed them up. The specific functions we wrote are:
\begin{itemize}
\item Data standardization (z-score)  
\item Compute the trace of $A^{T}A$
\item Add a scalar value to the diagonal of a matrix
\item Compute factor matrix \mats{F}{i} based on \eqref{eq:htfa_factor}
\item Compute residual errors $\normfro{\mats{X}{i} - \mats{W}{i}\mats{F}{i}}^2$ in \eqref{eq:htfa_subject_update_rbfs}
\end{itemize}

For HTFA we found that memory usage was increasing after every iteration, so we force garbage collection at the end of each local step using the Python \texttt{gc} module.

\subsection{Distributed Implementation}

We use MPI to enable multi-node parallelism. Subjects are distributed across MPI processes and we rely on parallelism within matrix operations to leverage multiple cores within a process. Communication between subjects maps directly to optimized MPI collective operations such as Reduce, Gather, and Bcast. We use these collectives exclusively for inter-subject communication. We found that having multiple MPI processes (i.e., multiple subjects) per node was using cores more efficiently than having fewer MPI processes and more shared-memory parallelism within a subject. As a result, we generally pack more subjects onto nodes until memory limitations arise. 

In order to enable parallel I/O from distributed filesystems, we store each subject's data in different files using NumPy's standard binary format. With this approach, each MPI process loads its subject's data in parallel with other processes.

\section{Experimental Setup}
\label{sec:experiments}

In this section, we describe the experimental setup, the real fMRI datasets we used for experiments, as well as the algorithm we designed to generate large-scale synthetic fMRI data based on the real fMRI datasets.

\subsection{Configuration}

We run our experiments on the Cori Phase I machine at the National Energy Research Scientific Computing Center (NERSC)~\cite{corinersc}. Cori Phase I is a Cray XC40\footnote{Software and workloads used in performance tests may have been optimized for performance only on Intel microprocessors. Performance tests, such as SYSmark and MobileMark, are measured using specific computer systems, components, software, operations and functions. Any change to any of those factors may cause results to vary. You should consult other information and performance tests to assist you in fully evaluating your contemplated purchases, including the performance of that product when combined with other products. For more information go to http://www.intel.com/performance}, currently with 1630 compute nodes and a Cray Aries interconnect providing 5.6 TB global bandwidth. Each compute node contains two 16-core Intel Xeon\footnote{Intel and Xeon are trademarks of Intel corporation in the U.S. and/or other countries.} E5-2698~v3 CPUs running at 2.3~GHz, and 128~GB DDR4 2133~MHz memory. Cori Phase I also has a distributed Lustre filesystem with 130 router nodes and 248 I/O servers, capable of providing 744~GB/s in bandwidth. We use the Anaconda distribution of Python version 2.7.11 with the following packages: MKL 11.3 Update 1, SciPy 0.17.0, and NumPy 1.10.4. We use mpi4py version 2.0.0, compiled with Cray-provided wrappers.

For HTFA, we sample voxels and TRs as described in Section~\ref{sec:htfamap} using random sampling with replacement. To select the number of subsamples, we use the mean between a percent of the data (25\% for voxels, 10\% for TRs) and a maximum number of samples (3000 for voxels, 300 for TRs). To estimate D-SRM performance in Gflop/s, we count the number of floating point operations used by the four main computations in D-SRM. For each subject and iteration, there are two $V_i \times T_i \times K$ matrix multiplies, one $V_i \times K \times K$ matrix multiply, and one $V_i \times K$ economy SVD operation. For a lower bound estimate, we assume that the SVD performs the same number of floating point operations as Householder QR ($2mn^2-\frac{2}{3}n^3$).

\subsection{Datasets}

We consider the following real neuroimaging datasets:

\begin{itemize}

\item \dataset{raider}: subjects received two different stimuli while in the fMRI scanner: the film ``Raiders of the Lost Ark'' (110 min) and a set of still images (7 categories, 8 runs)~\cite{Hanke2014forrest}. 

\item \dataset{forrest}: subjects listened to an audio version of the film ``Forrest Gump'' (120 min)~\cite{Hanke2014forrest}.

\item \dataset{greeneyes}: subjects listened to an ambiguous story (15 min)~\cite{Yeshurun2014greeneyes}.

\end{itemize}

An overview of the datasets is shown in Table~\ref{tab:realdatasets}. 
All datasets have the same number of voxels per subject, except \dataset{raider}. The datasets were stored in either float64 or float32 format, but all data is cast to float64 before processing. All data is stored in row-major order. 

\begin{table}
\caption{Real datasets}
\label{tab:realdatasets}
\centering
\begin{tabular}{l c r r r}
  \toprule
  Dataset & Subjects & Voxels & TRs & Size \\ 
  \midrule
  \dataset{raider} & 10 & $\sim$3,000 & 2201 & 658 MB\\
  \dataset{greeneyes} & 40 & 42,855 & 475 & 3.1 GB \\
  \dataset{forrest} & 18 & 629,620 & 3535 & 299 GB\\
  \bottomrule
\end{tabular}
\end{table}

We create a large synthetic dataset to test the weak scalability of D-SRM and HTFA D-MAP. We start from a real dataset and randomly permute the temporal order of the data for each voxel. To preserve the spatial structures of the brain, we spatially partition the voxels  and use the same permutation for all voxels in a partition.

The precise approach we use for synthetic data generation is described in Algorithm~\ref{alg:synth}. Each partition is filled with data from a different randomly-chosen seed subject. Next, the data in each partition is permuted across TRs, using the same permutation within each partition. We use \dataset{forrest} as seed dataset to generate a new 1024-subject synthetic dataset. We split each brain volume using $(16 \times 16 \times 8)$-voxel partitions. The resulting \dataset{forrest-1024} dataset is 17~TB in float64. 

\begin{algorithm}
\caption{Permutation-Based Synthetic Data Generation}
\label{alg:synth}
\begin{algorithmic}[1]
\REQUIRE \textbf{R}, a set of $N_R$ 4D (3D volume $\times$ \#TR) real datasets; \\$x, y, z$, the dimensions of the spatial partitions; \\$N_S$, the number of desired synthetic subjects.
\ENSURE \textbf{S}, a set of $N_S$ 4D synthetic datasets.
\FOR{\emph{i} in $1 \ldots N_{S}$}
\STATE{Seed random number generator with $i$}
\FOR{\textbf{each} spatial partition $(x, y, z)$}
\STATE $S_{i}(x, y, z) = R_{j}(x, y, z)$ for random $j$ in $1 \ldots N_R$.
\ENDFOR 
\FOR{each spatial partition $(x, y, z)$}
\STATE {Randomly permute $S_{i}(x, y, z)$ along TR dimension}
\ENDFOR 
\ENDFOR 
\end{algorithmic}
\end{algorithm}

\section{Results}
\label{sec:results}

We examine the performance improvements of our optimized algorithms compared to the baseline versions. We measure the impact of the number of factors on performance. We test strong scaling using real datasets and weak scaling using the synthetic dataset. Finally, we present scientific results obtained using our optimizations.

\subsection{Improvements compared to baseline implementations}

We demonstrate the performance advantage of our our improved SRM code using \dataset{raider} and our improved HTFA code using 20 subjects from \dataset{greeneyes}. The results are shown in Tables~\ref{tab:srm-single-node} and Table~\ref{tab:htfa-single-node} for a single node, and $K=60$. "Baseline HTFA" and "Baseline SRM" are our own implementations of the currently published algorithms. The "Baseline + Algorithmic Opt." performance includes only linear algebra transformations and algorithmic improvements, which are described in Sections \ref{sec:srm} and \ref{sec:htfa}. These optimizations achieved $671\times$ and $42\times$ speedups for SRM and HTFA respectively. Finally, D-SRM and HTFA D-MAP include code optimizations, such as custom parallel NumPy extensions, and across-subject MPI parallelism (20 MPI processes and 3 threads per process). These additional optimizations added $2.7\times$ and $2.3\times$ speedups for D-SRM and HTFA D-MAP, respectively.

\begin{table}
\caption{Single-node performance of SRM implementations on \dataset{raider}}
\label{tab:srm-single-node}
\centering
\begin{tabular}{l r r}
  \toprule
  Algorithm & Runtime (s) & Speedup \\ 
  \midrule
  Baseline SRM & 2655.25 & - \\
  Baseline SRM + Algorithmic Opt. & 3.95 & 671$\times$ \\
  D-SRM & 1.46 & 1812$\times$ \\
  \bottomrule
\end{tabular}
\end{table}

\begin{table}
\caption{Single-node performance of HTFA implementations on 20 subjects from \dataset{greeneyes}}
\label{tab:htfa-single-node}
\centering
\begin{tabular}{l r r}
  \toprule
  Algorithm & Runtime (s) & Speedup \\ 
  \midrule
  Baseline HTFA&  15162.96 & - \\
  Baseline HTFA + Algorithmic Opt.& 361.34 &  42$\times$ \\
  HTFA D-MAP & 153.84 &  99$\times$ \\
  \bottomrule
\end{tabular}
\end{table}

\subsection{Impact of number of factors on performance}
\begin{figure}
      \centering
      \includegraphics[trim=0 10 0 20,clip,width=0.7\linewidth]{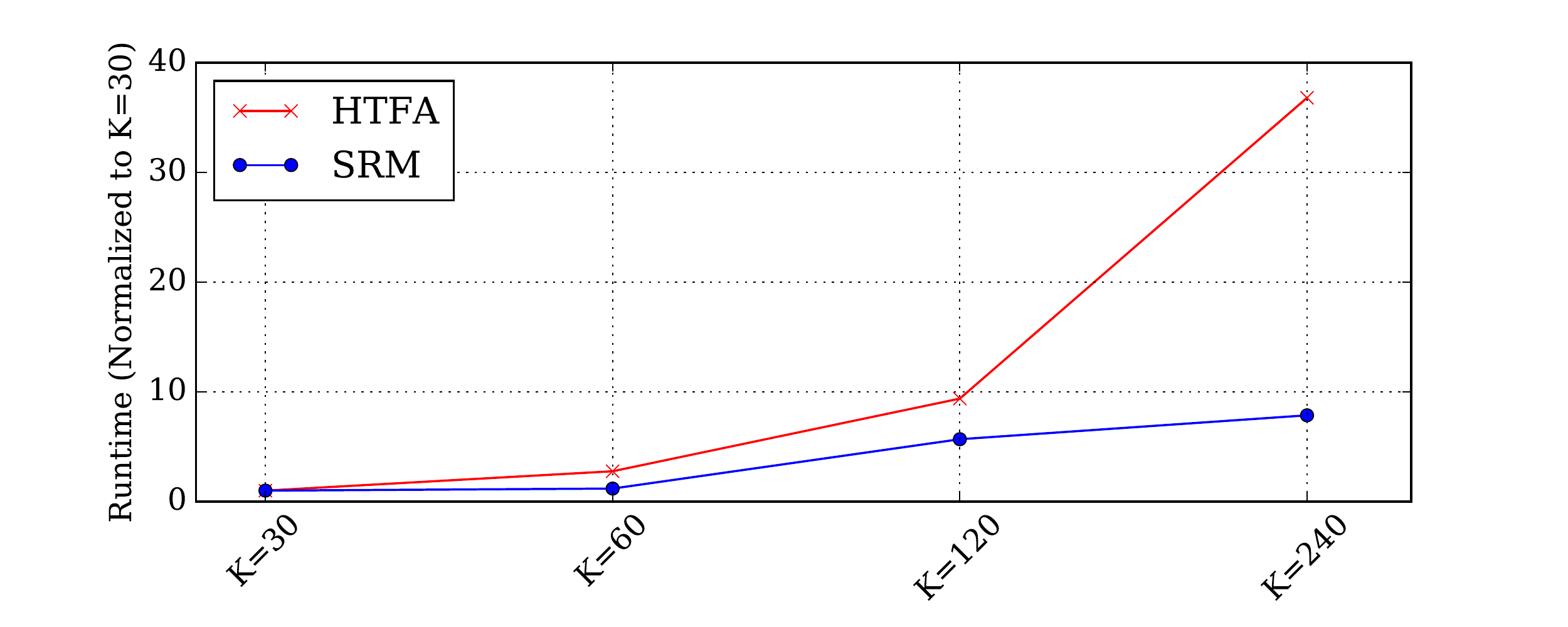} 
      \caption{Single-node runtime for different numbers of latent factors (K) normalized to the runtime for $K=30$, on 2 subjects from \dataset{greeneyes}, using 2 processes and 16 threads).} 
      \label{fig:k_scaling_greeneyes}
\end{figure}

Figure~\ref{fig:k_scaling_greeneyes} shows the effect of the number of latent factors on runtime. For D-SRM, the runtime is linear in $K$ for certain operations, such as two of the three matrix multiplications, and quadratic for other operations such as the SVD. For HTFA D-MAP, the runtime is linear in $K$ for the global template updates. The computation of the  Jacobian matrix depends on $K^2$, the number of voxels and the number of TRs. Overall, for both algorithms we observed a super linear increase of runtime as $K$ grows and a relatively large increase for HTFA D-MAP. For the remainder of the paper we fix $K$ to be 60, which was chosen based on prior neuroscience experiments using these algorithms~\cite{chen2015srm,manning2014topographic}.

\subsection{Strong scaling}

\begin{figure}
      \centering
      \includegraphics[trim=0 20 0 20,clip,width=0.7\linewidth]{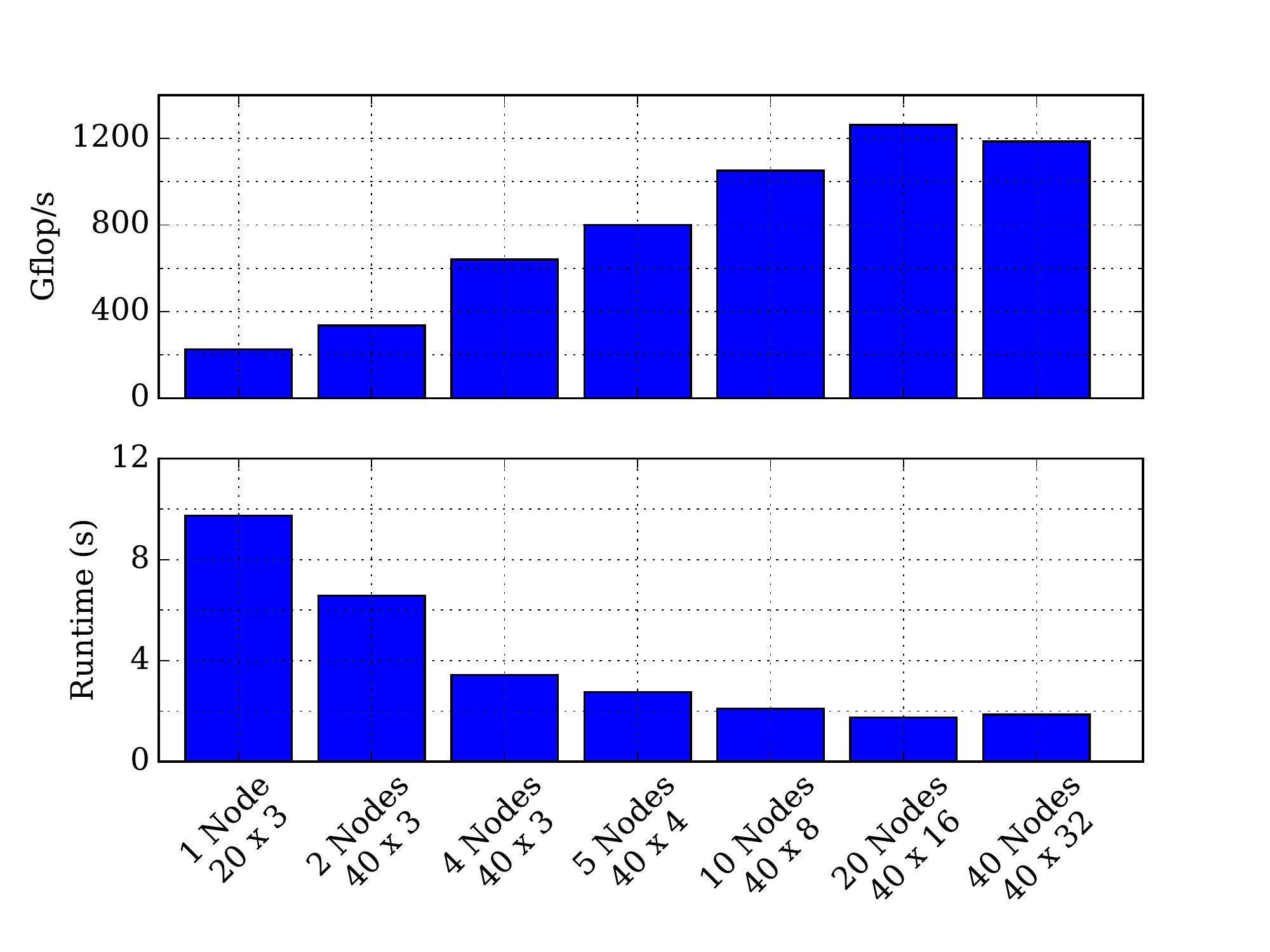} 
      \caption{D-SRM strong scaling on \dataset{greeneyes} up to 40 nodes with varying numbers of processes and threads (\#processes $\times$ \#threads). } 
      \label{fig:srm_greeneyes_strong_scaling}
\end{figure}

\begin{figure}
      \centering
      \includegraphics[width=0.7\linewidth]{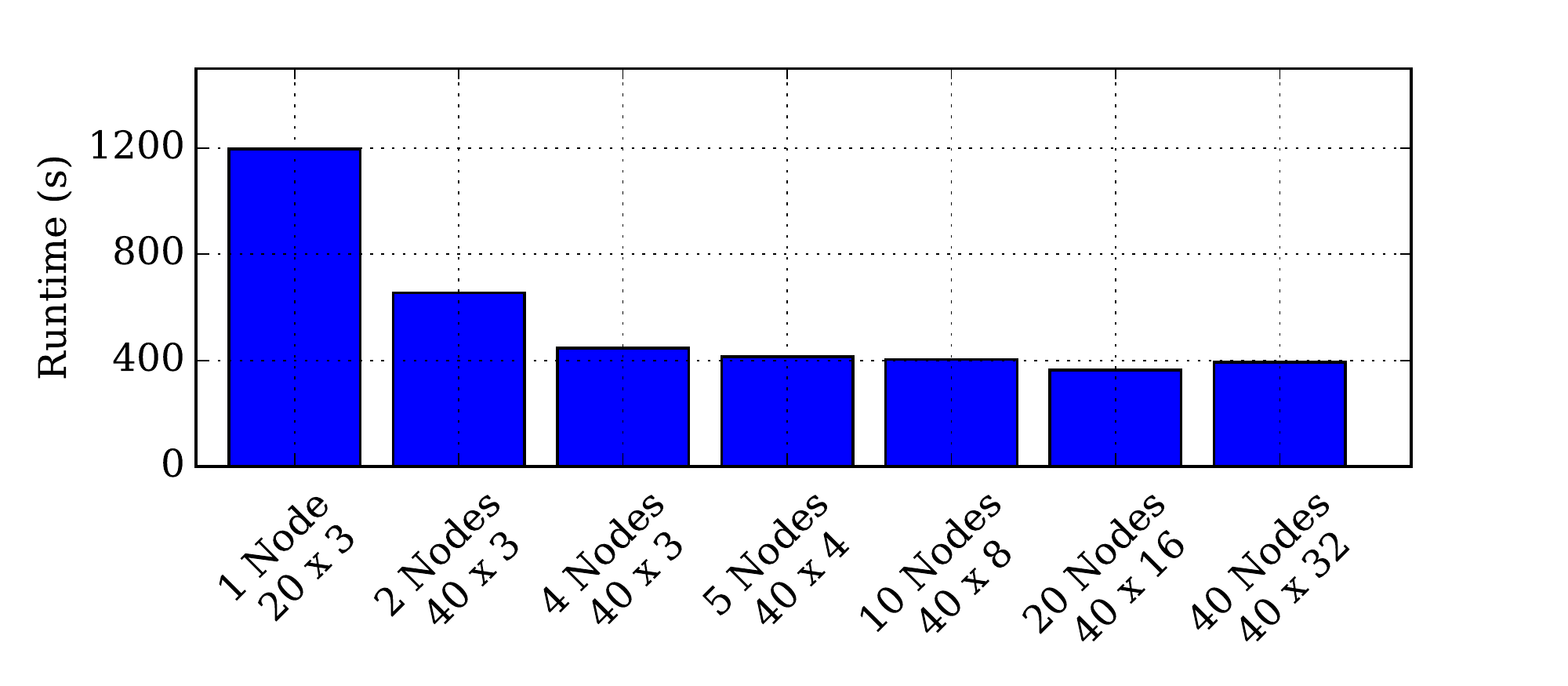} 
      \caption{HTFA D-MAP strong scaling on \dataset{greeneyes} up to 40 nodes with varying numbers of processes and threads (\#processes $\times$ \#threads). } 
      \label{fig:htfa_strong_greeneyes}
\end{figure}

Figures~\ref{fig:srm_greeneyes_strong_scaling} and~\ref{fig:htfa_strong_greeneyes} show strong scaling performance on \dataset{greeneyes}. We measure compute time of 10 iterations on setups ranging from 1 node with 32 cores to 40 nodes with 1280 cores in total. Note that the number of MPI processes we use must divide the number of subjects in the real dataset, 40.


For both D-SRM and HTFA D-MAP, we found that parallelizing across subjects using MPI was generally more effective in utilizing cores than relying on OpenMP parallelism to saturate the cores. This is only a consideration for datasets in which multiple subjects' data can fit in one node's memory. We experimented with different numbers of processes and threads and chose those that gave good performance. For example, on a single node, we use 20 processes with 3 threads per process (using Hyper-Threading), just like in the baseline comparison. For D-SRM, this configuration achieved 225 double precision Gflop/s out of a theoretical peak of 1177.6 Gflop/s. For both applications, we saw improved performance from 1 to 20 nodes, however the scaling slowed significantly after 5 nodes.  We attribute this behavior to (1) matrix operations having a tall-skinny shape and therefore not scaling as well as more compute-bound square matrix operations, and (2) increasing parallel overheads for larger node counts on the relatively small \dataset{greeneyes} dataset. Note that performance drops when using 1 process per node with 32 threads instead of 2 processes with 16 threads, because of parallel overhead and NUMA effects.

\subsection{Weak scaling}

\begin{figure}
      \centering
      \includegraphics[trim=0 10 0 20,clip,width=0.7\linewidth]{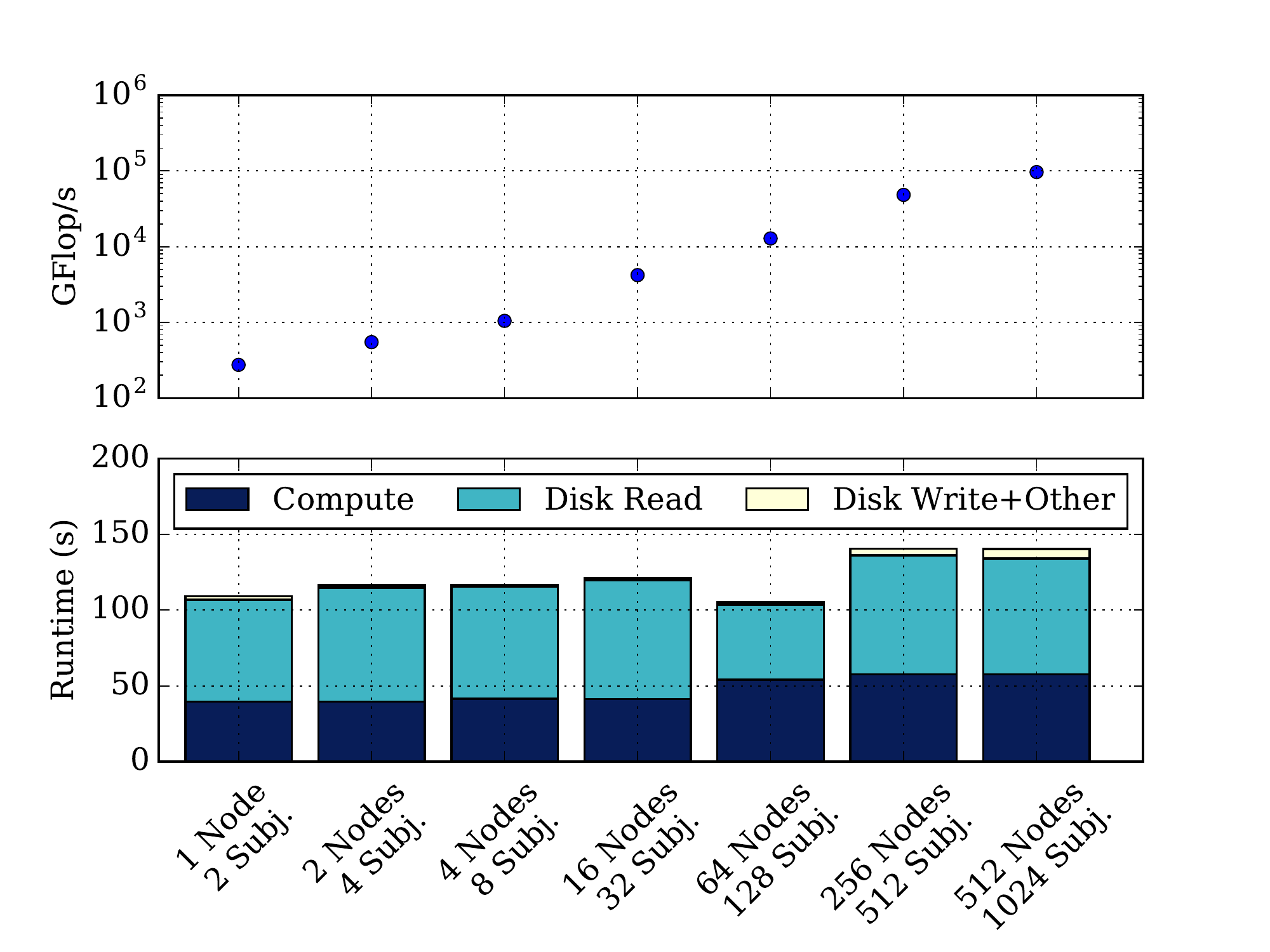} 
      \caption{D-SRM weak scaling on \dataset{forrest-1024} up to 512 nodes, broken down into disk I/O and compute time. Gflop/s numbers consider only compute time, without I/O. Log axis for Gflop/s.} 
      \label{fig:srm_forrest_weak_scaling}
\end{figure}

We test D-SRM and HTFA D-MAP weak scaling on \dataset{forrest-1024}. For D-SRM, we processes two subjects per node, and for HTFA D-MAP we process one subject per node, which is driven by memory limitations. The results in Figure~\ref{fig:srm_forrest_weak_scaling} show that compute time per subject (excluding I/O) rises by only $1.46\times$ for D-SRM from 1 node to 512 nodes and $1.07\times$ for HTFA D-MAP from 1 node to 1024 nodes. Disk I/O takes a significantly larger portion of total time for D-SRM compared to HTFA D-MAP, because D-SRM spends relatively less time computing than HTFA D-MAP while having to load the same amount of data. 

\begin{figure}
      \centering
      \includegraphics[width=0.7\linewidth]{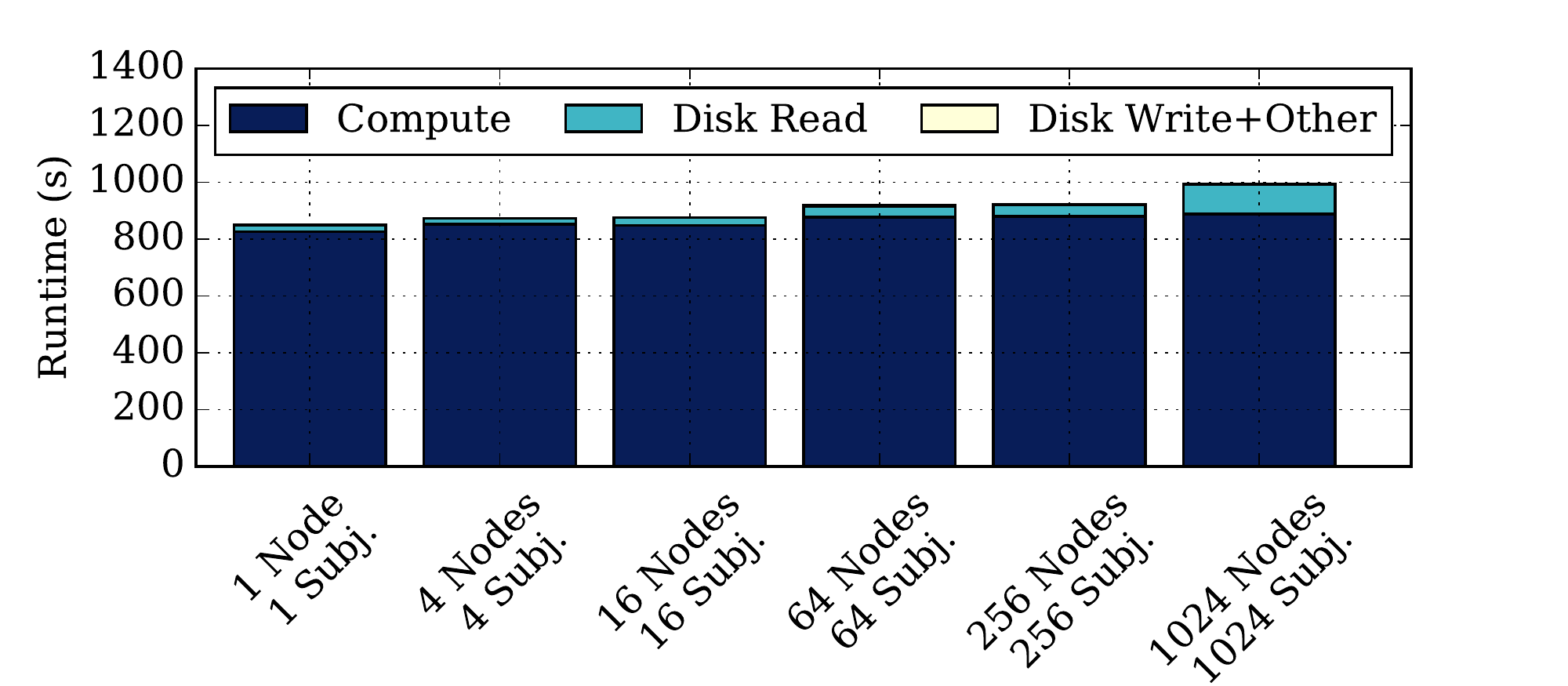} 
      \caption{HTFA D-MAP weak scaling on \dataset{forrest-1024} up to 1024 nodes, broken down into disk I/O and compute time.} 
      \label{fig:htfa_weak_forrest}
\end{figure}



%

%

\section{Discussion}
\label{sec:discussion}

The computation and memory requirements of multi-subject analysis techniques, such as SRM and HTFA, are largely driven by the dimensionality of the data. Namely, neuroimaging data tends to have extremely high dimensionality per sample (e.g., 600,000 voxels), compared to relatively few samples (e.g., 3,500 TRs). This extreme skew in the data dimension makes linear algebra transformations and algorithmic optimizations potentially very profitable, because we have the potential to change the computational complexity from something that depends on the number of voxels, to something that depends only on the number of samples or latent factors. The matrix inversion lemma does this, and we applied it to both SRM and HTFA. The shape of the data also means that our optimized implementations mainly operated on so-called "tall-skinny" matrices, where the number of rows is much larger than the number of columns. This had a direct impact on our performance and our ability to scale to multiple cores within a node. 

The computational structure of the multi-subject analysis techniques we studied was very amenable to multi-node parallelism. This also follows directly from the nature of the models we are fitting. Both models allow for a global (across-subject) representation of neural activity, and a local (per-subject) configuration. The algorithms to fit the models will alternate between fitting local models with the global model fixed, and vice versa. This translates directly to map-reduce style parallelism. In the case of HTFA and SRM, fitting the global model was relatively inexpensive compared to fitting local models, which helped scalability. However, other multi-subject model-fitting algorithms may not have this feature.






\section{Conclusion}
\label{sec:conclusion}


We have introduced highly-parallel codes for two applications of factor analysis on multi-subject neuroimaging datasets, SRM and HTFA.  We demonstrated good single-node performance and scaling on a state-of-the-art supercomputer, up to 32,768 cores. This makes the discovery of latent factors present in large subject populations practical, allowing subject-specific and shared responses to stimuli to be readily identified and interpreted. To achieve these results, we created highly-optimized implementations of the factor analysis algorithms starting from validated baseline implementations.  
For D-SRM, we achieved a $1,812\times$ speedup over the baseline code by deriving an alternate formula for the large matrix inversion in the critical E-step of the EM algorithm, reducing distributed communications during the M-step through changes to hyperparameter computation, an d explicit node-level parallelization with MPI. 
For HTFA D-MAP, we achieved 99$\times$ speedup over the baseline on a medium size dataset by splitting the variables of the non-linear least squares problem to reduce the size of the Jacobian matrix and converge faster, using cache-based table lookups for distance calculations, reducing the number of matrix inversions, and explicit garbage collection.  Our experience with these two applications allowed us to identify several key performance engineering recommendations that apply to factor analysis of multi-subject neuroimaging data in general.

We expect that our work will enable neuroscientists to develop population-based denoising techniques, design studies to better understand similarities and differences in cognitive function within populations, and perform more sophisticated studies of human interaction. To this end, we are releasing our code as open-source software on GitHub to allow 
the use our optimized versions of these algorithms on a supercomputer while retaining the programming benefits of Python.





\bibliographystyle{IEEEtran}
\bibliography{IEEEabrv,biblio,refs_srm}

\end{document}